%% file: main.tex
\documentclass[10pt,twocolumn,letterpaper]{article}

\usepackage{cvpr}              

\usepackage[accsupp]{axessibility}
\usepackage{graphicx}
\usepackage{amsmath}
\usepackage{amssymb}
\usepackage{booktabs}

\usepackage[pagebackref,breaklinks,colorlinks]{hyperref}

\usepackage[capitalize]{cleveref}
\crefname{section}{Sec.}{Secs.}
\Crefname{section}{Section}{Sections}
\Crefname{table}{Table}{Tables}
\crefname{table}{Tab.}{Tabs.}

\def\cvprPaperID{4032} 
\def\confName{CVPR}
\def\confYear{2022}

\begin{document}

\title{PanopticDepth: A Unified Framework for Depth-aware Panoptic Segmentation}

\author{
Naiyu~Gao$^{1,2}$,
Fei~He$^{1,2}$,
Jian~Jia$^{1,2}$,
Yanhu~Shan$^{4}$,
Haoyang~Zhang$^{4}$,
Xin~Zhao$^{1,2}$\thanks{Corresponding author},
Kaiqi~Huang$^{1,2,3}$
\\
$^{1}$
CRISE, Institute of Automation, Chinese Academy of Sciences\\
$^{2}$
School of Artificial Intelligence, University of Chinese Academy of Sciences\\
$^{3}$
CAS Center for Excellence in Brain Science and Intelligence Technology,
$^{4}$
Horizon Robotics, Inc.\\
{\tt\small \{gaonaiyu2017,hefei2018,jiajian2018\}@ia.ac.cn,\{xzhao,kaiqi.huang\}@nlpr.ia.ac.cn}\\
{\tt\small\{yanhu.shan,haoyang.zhang\}@horizon.ai
}}
\maketitle

\begin{abstract}
This paper presents a unified framework for depth-aware panoptic segmentation (DPS), which aims to reconstruct 3D scene with instance-level semantics from one single image. Prior works address this problem by simply adding a dense depth regression head to panoptic segmentation (PS) networks, resulting in two independent task branches. This neglects the mutually-beneficial relations between these two tasks, thus failing to exploit handy instance-level semantic cues to boost depth accuracy while also producing sub-optimal depth maps. To overcome these limitations, we propose a unified framework for the DPS task by applying a dynamic convolution technique to both the PS and depth prediction tasks. Specifically, instead of predicting depth for all pixels at a time, we generate instance-specific kernels to predict depth and segmentation masks for each instance. Moreover, leveraging the instance-wise depth estimation scheme, we add additional instance-level depth cues to assist with supervising the depth learning via a new depth loss. Extensive experiments on Cityscapes-DPS and SemKITTI-DPS show the effectiveness and promise of our method. We hope our unified solution to DPS can lead a new paradigm in this area. Code is available at \url{https://github.com/NaiyuGao/PanopticDepth}.
\end{abstract}

\input{Sec1_introduction.tex}
\input{Sec2_related_work.tex}
\input{Sec3_method.tex}
\input{Sec4_experiments.tex}

\section{Conclusion}
\label{sec:concl}
In this paper, a unified framework for depth-aware panoptic segmentation is proposed by predicting the mask and depth values for each thing/stuff in the same instance-wise manner. High-level object information is introduced into the depth estimation by adopting the dynamic kernel technique. Moreover, each instance depth map is normalized with a depth shift and a depth range to ease the learning of shared depth embedding. Furthermore, a new depth loss is proposed to supervise the depth learning with instance-level depth cues. Experiments on Cityscapes-DPS and SemKITTI-DPS benchmarks demonstrate the effectiveness of the proposed approach. We hope our unified solution to depth-aware panoptic segmentation can lead a new paradigm in this area.

\section*{Acknowledgment}
This work is supported in part by the National Natural Science Foundation of China (Grant No.61721004 and Grant No.61876181), the Projects of Chinese Academy of Science (Grant No. QYZDB-SSW-JSC006), the Strategic Priority Research Program of Chinese Academy of Sciences (Grant No. XDA27000000), and the Youth Innovation Promotion Association CAS.

{\small
\bibliographystyle{unsrt}
\bibliography{egbib}
}

\end{document}

%% file: Sec1_introduction.tex
\begin{figure}
    \center
    \includegraphics[width=1.\linewidth]{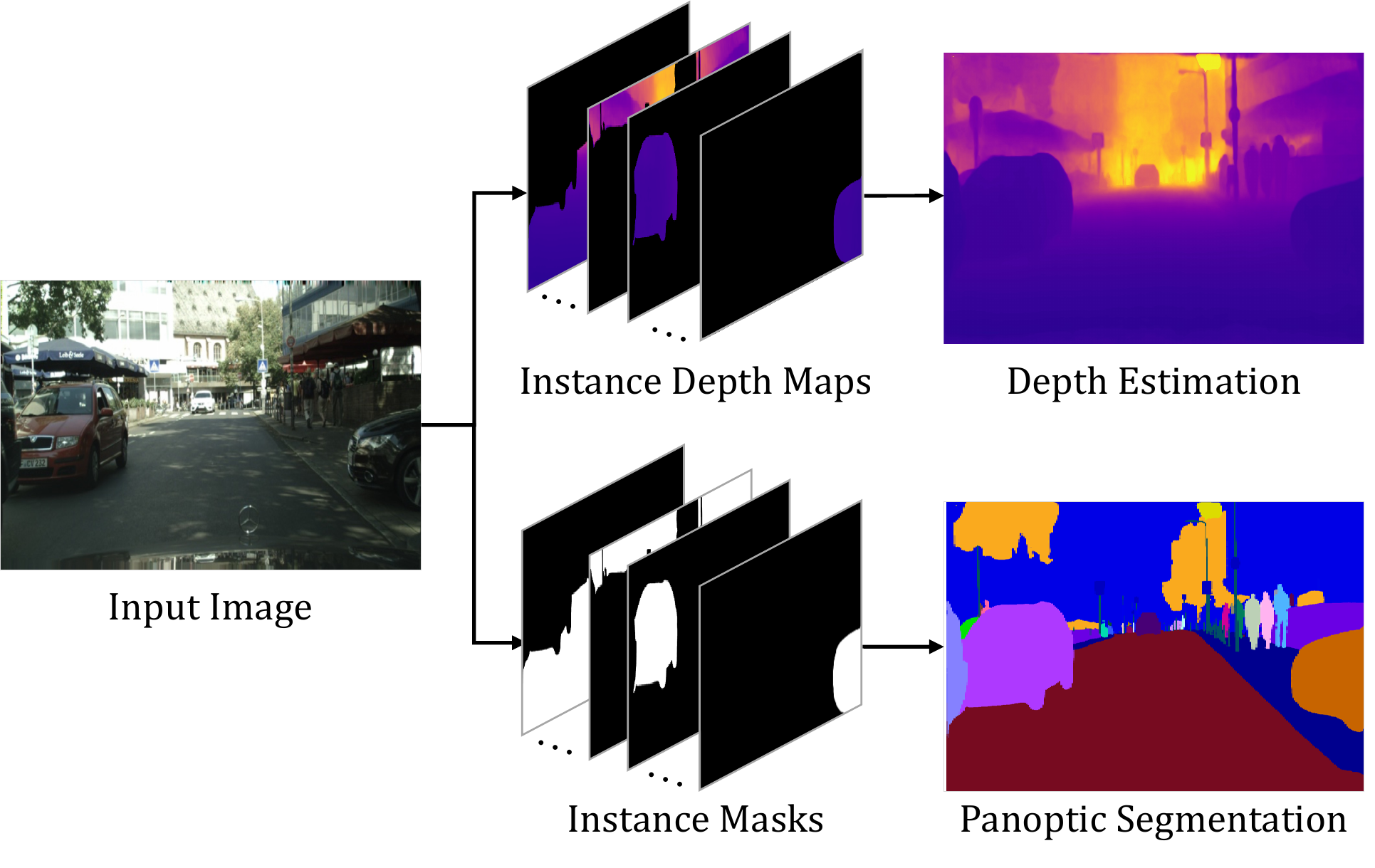}
    \caption{Illustration of our unified solution to depth-aware panoptic segmentation which requires assigning each pixel in a single image a depth value, a semantic class label and an instance ID. Instead of predicting pixel-wise depth, we predict instance-wise depth by instance-specific convolution kernels, which shares the same manner of instance mask generation.}
  \label{fig:overview}
\end{figure}

\section{Introduction}
\label{sec:intro}
Depth-aware panoptic segmentation (DPS) is a new challenging task in scene understanding, attempting to build 3D scene with instance-level semantic understanding from a single image. Its goal is to assign each pixel a depth value, a semantic class label and an instance ID. Thus, solving this problem involves monocular depth estimation~\cite{fu2018deep} and panoptic segmentation~\cite{panoptic}. Naturally, a straightforward solution to DPS is to add a dense depth regression head to panoptic segmentation (PS) networks~\cite{qiao2021vip,liu2021towards,schon2021mgnet}, producing a depth value for each labelled pixel.

This method is intuitive yet sub-optimal. Since it tackles the two tasks with two independent branches, it does not explore the mutually-beneficial relations between them, especially failing to exploit handy instance-level semantic cues to boost depth accuracy. We observe that pixels of adjacent instances generally have discontinuous depth. For examples, two vehicles in a line are likely to have different depth. Therefore, it is hard to predict accurate depth for both vehicles using the same pixel-wise depth regressor. 
On the other hand, as indicated by previous works~\cite{zhang2018joint,xu2018pad,meng2019signet,jiao2018look}, considering that these pixels are of different vehicles, it benefits the depth estimation if separate regressors are used respectively.

Following the train of thought above, we propose PanopticDepth in this paper, a unified model that predicts the mask and depth values in the same instance-wise manner~(Figure~\ref{fig:overview}). In contrast to predicting depth values for all pixels at a time, we manage to estimate depth for each thing/stuff instance, which also shares the way of generating instance masks. To this end, we employ the technique of dynamic convolution~\cite{tian2020conditional,wang2020solov2,li2021fully,cheng2021maskformer} in producing both instance masks and depth, which unifies the pipelines of mask generation and depth estimation.

Specifically,  we first generate instance-specific mask and depth kernels for each instance concurrently, which is similar to the method used in~\cite{li2021fully}. Then, we apply the mask kernels to the mask embeddings and the depth kernels to the depth embeddings, producing the mask and depth map for each instance, respectively. Finally, we merge individual instance masks into a panoptic segmentation map, following a similar process presented in~\cite{panopticFPN}. According to the panoptic segmentation results, we then aggregate each instance's depth into a whole depth map. As a result, we get both the panoptic segmentation and depth map for one image. Figure~\ref{fig:pipeline} shows the pipeline.

Our method unifies the depth estimation and panoptic segmentation approaches via an instance-specific convolution kernel technique, which also in turn improves the performance on both tasks (Table~\ref{tab:dps_ablation} and Table~\ref{tab:kitti}). Thanks to the dynamic convolution kernel technique~\cite{li2021fully}, the learned instance depth regressor aggregates not only global context, but local information, such as instance shapes, scales and positions, into instance depth prediction. Such information turns out essential to obtain accurate depth values especially those at instance boundaries (Figure~\ref{fig:vis_boundary}).

Furthermore, in order to ease the depth estimation, inspired by Batch Normalization~\cite{ioffe2015batch}, we propose to represent each instance depth map as a triplet, \ie a normalized depth map, a depth range and a depth shift. In general, depth values of different instances may vary greatly, like a long vehicle with a length of 70m v.s. a small car with a length of 4.5m. This large variation in scale may cause difficulties in learning shared depth embedding. To tackle this problem, we propose to represent the depth map with the aforementioned triplet and normalize the values of original instance depth map to [0, 1]. This improves the learning effectiveness (Table~\ref{tab:dps_ablation} and Table~\ref{tab:kitti}). At the same time, in addition to the traditional pixel-level depth supervision, we add instance-level depth statistics based on the new depth map representation, \eg the depth shift, to reinforce the depth supervision. We also propose a corresponding depth loss to accommodate this new supervision~(Sec:\ref{subsubsec:depth_loss}), which helps improve the depth prediction.

Through extensive experiments on Cityscapes-DPS~\cite{qiao2021vip} and SemKITTI-DPS~\cite{qiao2021vip}, we demonstrate the effectiveness of our unified solution to the depth-aware panoptic segmentation. We hope our unified framework can lead a new paradigm in this challenging task.

%% file: Sec2_related_work.tex
\section{Related Work}
\label{sec:rela}
\subsection{Panoptic Segmentation}
Panoptic segmentation~\cite{panoptic} (PS) is a recently proposed vision task, which requires generating a coherent scene segmentation for an image. It unifies the tasks of semantic segmentation and instance segmentation. Early approaches~\cite{Liu_2019_CVPR,Li_2019_CVPR,panopticFPN,xiong19upsnet,porzi2019seamless,mohan2021efficientps,li2018learning,Li_2020_CVPR} to PS mainly rely on individual semantic segmentation models~\cite{zhao2017pyramid,chen2017rethinking,chen2018deeplabv2} to get pixel-level segmentation for stuff classes and two-stage instance segmentation models~\cite{He_2017_ICCV} to get instance-level segmentation for things, and then fuse them~\cite{panoptic,Lazarow_2020_CVPR} to get the final panoptic segmentation.
Instead of adopting two-stage instance segmentation models, recent works\cite{real-time-panoptic,hong2021lpsnet, arxiv_2019_yang_deeperlab,Liu_2018_ECCV,SSAP_Gao_ICCV,SSAP_Gao_CSVT,Sofiiuk_2019_ICCV,Neven_2019_CVPR} attempt to solve panoptic segmentation in a one-stage manner. Among them, DeeperLab~\cite{arxiv_2019_yang_deeperlab} proposes to solve panoptic segmentation by predicting key-point and multi-range offset heat-maps, followed by a grouping process~\cite{Papandreou_2018_ECCV}. Similarly, the series of Panoptic-DeepLab ~\cite{cheng2020panoptic,wang2020axial,chen2020scaling} predict instance centers as well as offsets from each pixel to its corresponding center.
More recently, inspired by the instance-specific kernel method used in~\cite{Bolya_2019_ICCV,tian2020conditional,wang2020solo,wang2020solov2}, some works~\cite{li2021fully,zhang2021k,cheng2021maskformer,gao2021learning,wang2021max} start to apply this technique in PS, achieving unified modelling of both stuff and thing segmentation.

In this work, we extend the PS task by predicting additional depth for each instance, to build a 3D semantic scene. Similar to~\cite{li2021fully,zhang2021k,cheng2021maskformer,gao2021learning,wang2021max}, our proposed unified framework for this problem is also inspired by the instance-specific kernel technique, especially the work~\cite{li2021fully}. 

\subsection{Monocular Depth Estimation}
Monocular depth estimation aims to predict depth from a single image. Before deep learning era, there have already been many attempts~\cite{hoiem2005geometric,saxena2008make3d,liu2010single,liu2014discrete}. Recently, a variety of better-performing methods based on deep networks have been proposed. Laina~\etal~\cite{laina2016deeper} propose a fully convolutional architecture with up-projection blocks for high-dimensional depth regression, while Li~\etal~\cite{li2017two} design a two-streamed network to predict fine-scaled depth. Besides supervised learning, other learning techniques have also been explored to improve depth estimation generalization, including self-supervised learning~\cite{garg2016unsupervised,godard2019digging,godard2017unsupervised,wang2018learning,zhou2017unsupervised}, transfer learning with synthetic images~\cite{atapour2018real,zheng2018t2net}, and learning relative depth orders from web data for depth perception~\cite{chen2016single,li2019learning,li2018megadepth,wang2019web,wang2018deeplens,xian2018monocular}.

Most aforementioned works predict depth from pixel perspective, in contrast, our method predicts depth for each instance, which incorporates instance-level cues into depth estimation.

\vspace{-4mm}
\paragraph{Enhance depth with semantic segmentation. }
There are works~\cite{zhang2018joint,xu2018pad,meng2019signet,eigen2015predicting,mousavian2016joint,wang2015towards} that attempt to improve monocular depth estimation with semantic segmentation. 
Liu~\etal~\cite{liu2010single} propose to boost depth estimation with semantic labels by Markov random fields. 
Pad-Net~\cite{xu2018pad} leverages intermediate depth and segmentation predictions to refine the final output. 
Jiao~\etal~\cite{jiao2018look} refine depth predictions with attention-driven loss or task-level interactions. 
SigNet~\cite{meng2019signet} integrates semantic segmentation with instance edges to model depth estimation in an unsupervised manner.

These works show the mutual benefit of simultaneously conducting depth estimation and semantic segmentation, but we argue that we can achieve better results if we perform instance-wise depth estimation with panoptic segmentation, since we now have instance-level cues to facilitate depth estimation.

\vspace{-4mm}
\paragraph{Enhance depth with panoptic segmentation. }
More recently, some works~\cite{wang2020sdc,schon2021mgnet,saeedan2021boosting,qiao2021vip} have started to explore the joint learning of monocular depth estimation and panoptic segmentation, but most of them utilize panoptic segmentation to enhance depth just by sharing backbone features~\cite{qiao2021vip} or adding constraints\cite{saeedan2021boosting,schon2021mgnet}. By contrast, we unify the panoptic segmentation and instance-wise depth estimation in one framework through the instance-specific convolution technique.

Among them, SDC-Depth~\cite{wang2020sdc} also explores the idea of instance-wise depth estimation, but it is different from our work in many aspects.
First of all, our target task is different from its. The goal of SDC-Depth is just to estimate depth where instance masks is used only to provide clues for improving the depth map. In contrast, our DPS task requires producing both high-quality masks and depth maps, which is more challenging.
Besides, we adopt a different instance depth normalization scheme, which allows us to supervise learning depth estimation with both pixel- and instance-level cues at the meantime.
Finally, our approach is built upon a dynamic-convolution-based panoptic segmentation model~\cite{li2021fully}, which generates more precise segment boundaries, rather than the two-stage approach employed in SDC-Depth.

%% file: Sec3_method.tex
\section{PanopticDepth}
\label{sec:meth}

\begin{figure*}
  \center
  \includegraphics[height=88mm]{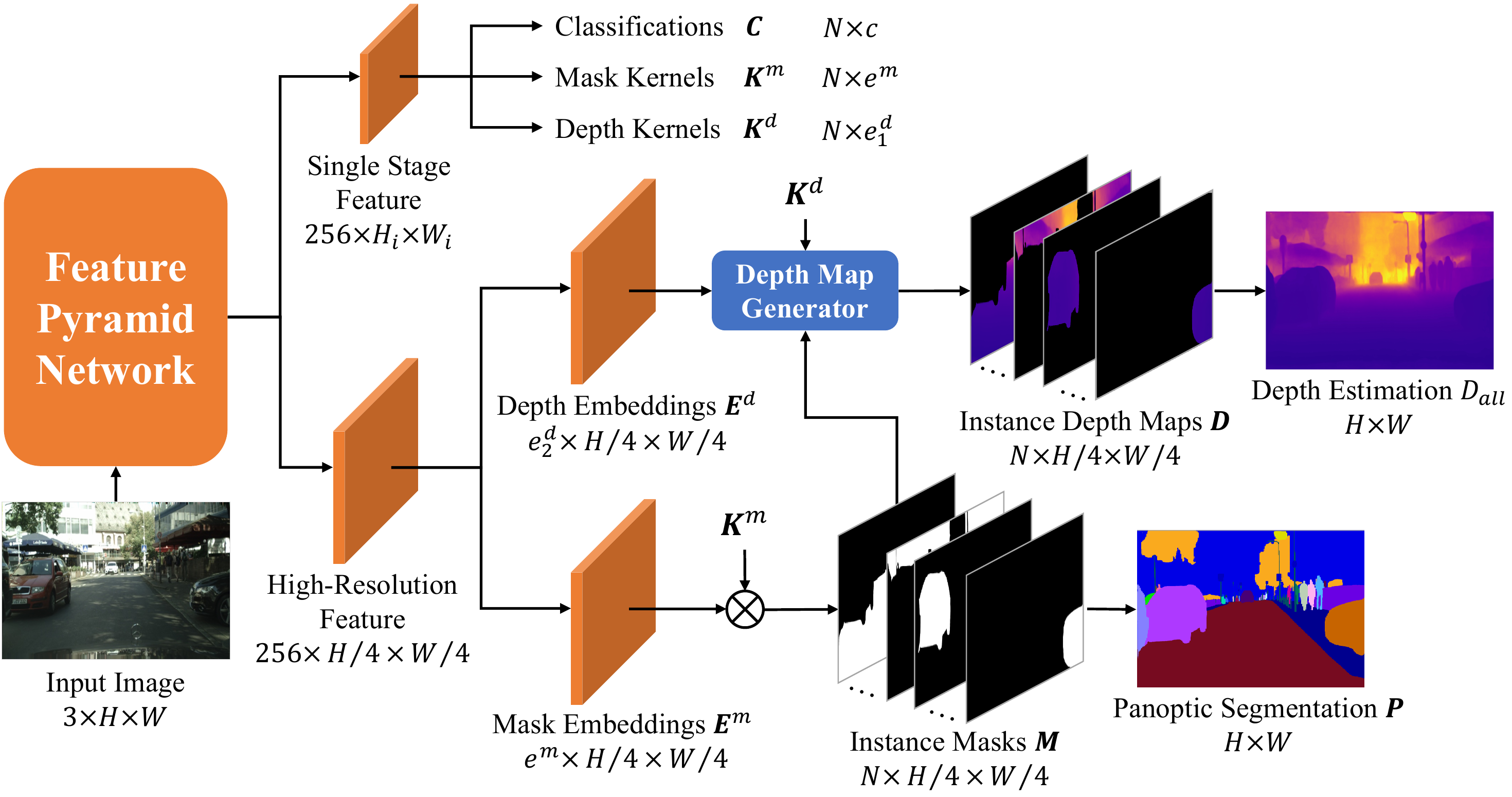}
  \caption{The framework of our proposed PanopticDepth. 
	$H$ and $W$ are the height and width of the input image respectively. 
	$N$ denotes the number of thing and stuff instances.
	$c$ is the number of categories.
	$e^m$, $e^d_1$, and $e^d_2$ are embedding dimensions of mask kernels, depth kernels and the depth embedding map, respectively.
	$\otimes$ represents convolution.
	}
  \label{fig:pipeline}
\end{figure*}

We propose PanopticDepth, a unified model for depth-aware panoptic segmentation, which predicts the mask and depth values in the same instance-wise manner. Apart from backbone and the feature pyramid network~\cite{lin2017feature}, it consists of three main sub-networks, including a kernel producer for generating instance classification, instance-specific mask and depth convolution kernels, a panoptic segmentation model for generating instance masks, and an instance-wise depth map generator for estimating instance depths. The architecture of our networks is shown in Figure~\ref{fig:pipeline}. We elaborate each module in the following sections.

\subsection{Kernel Producer}
In our system, we first generate instance classification, mask convolution kernels and depth estimation kernels by a kernel producer sub-network (upper part of Figure~\ref{fig:pipeline}). Our kernel producer is built on the state-of-the-art panoptic segmentation model, PanopticFCN~\cite{li2021fully}, which adopts a dynamic convolution technique for PS but requires less training time and GPU memory compared to other recent methods~\cite{cheng2021maskformer,zhang2021k}. We briefly describe the process of kernel generation here. For more details please refer to~\cite{li2021fully}.

Given multiple stage FPN features $X$, our kernel producer outputs $N$ instance classifications, $N$ mask kernels, and $N$ depth kernels. It has two stages, that is, kernel generator and kernel fusion. 

At the kernel generator stage, taking as input a single stage feature $X_{i}$ from the i-th stage in FPN, the generator produces a kernel weight map $G_{i}$, and two position maps for things $L_{i}^{th}$ and stuff $L_{i}^{st}$ respectively, where each thing is represented by an object center and stuff by a region.

Given the position maps and kernel weight map from each FPN stage, at the kernel fusion stage, we merge repetitive kernel weights from multiple FPN stages. This is done through a proposed adaptive kernel fusion (AKF) operation, which is an improved fusion mechanism over the original average clustering used in PanopticFCN. Our adaptive fusion mechanism improves 1.2\% PQ without introducing extra parameters.
Specifically, to generate a mask kernel $K^{m}_{k}$ for the k-th instance, we compute it through:
\begin{equation}
 K^{m}_{k}=\frac{\sum(G*R_k)}{\sum R_k},\label{eq:akf}
\end{equation}
where $*$ denotes element-wise multiplication, and $R$ represents positions with peak scores in $L^{th}$ for things and positions with the highest score for corresponding categories in $L^{st}$ for stuff, respectively. 
Kernels with large cosine similarities are also fused, as done in ~\cite{li2021fully}.

For depth kernel generation, the similar process is performed. 
In this way, given FPN features, our kernel producer generates instance classification $C \in R^{N \times c}$, mask kernels $K^m \in R^{N \times e^{m}}$, and depth kernels $K^d \in R^{N \times e^d_1}$, where
$c$ is the number of categories,
$e^m$ and $e^d_1$ are dimensions of mask and depth kernels, respectively.

\subsection{Panoptic Segmentation}
\label{sec:baseline}
We adopt an instance-specific kernel method to perform panoptic segmentation~\cite{li2021fully}. As shown in the bottom part of Figure~\ref{fig:pipeline}, the masks $M$ of thing and stuff instances are derived by convolving a shared high-resolution mask embedding map $E^{m} \in R^{e^{m}\times H/4 \times W/4}$ with the mask kernels $K^{m}$, followed the Sigmoid activation:
\begin{equation}
M=\text{Sigmoid}(K^m \otimes E^m).
\end{equation}
Redundant instance masks are firstly discarded as done in ~\cite{panopticFPN}. After that, all remaining instance masks are merged with {\tt{argmax}} to yield the non-overlapped panoptic segmentation result, so that each pixel is assigned to a thing or stuff segment, with no pixel labeled as `VOID'.

Moreover, we propose an additional training process to bridge the performance gap between training and testing.
Concretely, we find distant instances are often fused by the learned model, which is caused by the widely employed image cropping strategy. Some previous works~\cite{cheng2020panoptic,qiao2021vip} tackle this problem by directly training models with original resolution images, which is effective but increases the GPU memory footprint dramatically.
In contrast, we propose a more efficient training strategy, \ie, fine-tuning the learned model at the full image scale, but with a small batch size. More detailed training process can be found in Subsection~\ref{sec:exp:imple}.

\subsection{Instance-wise Depth Estimation}
\label{sec:depth}
We predict depth for each instance through the same instance-specific kernel technique used in panoptic segmentation, which unifies the pipelines of depth estimation and panoptic segmentation. As shown in the middle part of Figure~\ref{fig:pipeline}, we first run the depth kernels on the depth embedding to generate instance depth maps, and then merge these individual maps in accordance with the panoptic segmentation results to yield the final whole depth map. In this section, we first introduce the depth map generator and then a new depth loss.

\subsubsection{Depth Map Generator}
\label{sec:depth_map_generator}
Given instance-specific depth kernels $K^d \in R^{N \times e^d_1}$ and the shared depth embedding $E^d\in R^{e_{2}^{d}\times H/4 \times W/4}$ (we set $e^d_1=e^d_2$), similar to the instance mask generation process, we first generate the normalized instance depth map $D^{'}$ through convolution and Sigmoid activation, and then unnormalize it to the depth map $D$ through Eq~\ref{eq:t1} or Eq~\ref{eq:t2}:
\begin{align}
&D^{'}=\text{Sigmoid}(K^d \otimes E^d),\\
&\mathcal{T}_1(D | D^{'}, d^r, d^s) = d_\text{max}\times(d^r \times D^{'} + d^s),
\label{eq:t1}\\
&\mathcal{T}_2(D | D^{'}, d^r, d^s) = d_\text{max}\times\big[d^r \times (D^{'} - 0.5)  + d^s\big],
\label{eq:t2}
\end{align}
where $d_\text{max}$ controls the depth scale and is set to 88 to be consistent with the depth range of Cityscapes-DPS and SemKITTI-DPS.

The reason why we normalize the depth map is that different instances have widely varied depth ranges, making it hard to learn effective shared depth embedding. To ease the learning of depth estimation and inspired by Batch Normalization~\cite{ioffe2015batch}, we predict a normalized depth map $D^{'}$  instead, which is obtained by normalizing the instance depth map $D$ with two predicted instance-level depth variables, depth range $d^{r}\in R^{N\times1}$ and depth shift $d^{s}\in R^{N\times1}$. They describe the depth bias and variance of each instance, respectively, and can be derived from high-level features in parallel with depth kernels by simply setting $e^d_1=e^d_2+2$. Note that $0\leq d^{s} \leq 1$ and $0\leq d^{r} \leq 1$ after Sigmoid activation.

In this way, the normalized depth map $D^{'}$ only encodes relative depth values within each instance, and thus can be more easily learned. Besides, we develop two normalization schemes, \ie Eq.~\ref{eq:t1} and Eq.~\ref{eq:t2}, and find that the latter one works better.

After getting all instance depth maps, we aggregate them into a whole image depth map, according to the non-overlapped panoptic segmentation masks $M$. This generates precise depth values at instance boundaries.

\subsubsection{Depth Loss}
\label{subsubsec:depth_loss}
Following~\cite{qiao2021vip}, we develop the depth loss function based on the combination of scale-invariant logarithmic error~\cite{eigen2014depth} and relative squared error~\cite{geiger2012we}, both of which are popular depth metrics and can be directly optimized end-to-end. Specifically, 
\begin{align}
\widetilde{L}_{dep}(\bf{d},\hat{\bf{d}}) =& \frac{1}{n}\sum_{j}(\log d_j-\log \hat{d}_j)^2 \nonumber \\
-& \frac{1}{n^2}\big(\sum_j\log d_j-\log \hat{d}_j\big)^2 \nonumber\\
+&\Big[\frac{1}{n}\sum_j(\frac{d_j-\hat{d}_j}{\hat{d}_j})^2\Big]^{0.5},
\end{align}
where $\bf{d}$ and $\hat{\bf{d}}$ denote the predicted and the ground-truth depth, respectively.

Thanks to the instance-wise depth estimation means, we can now learn the depth prediction under both the traditional pixel-level supervision and extra instance-level supervision, which empirically improves the depth accuracy. To enable the dual supervision, our final depth loss $L_{dep}$ includes two loss terms. One is the pixel-level depth loss $L^{P}_{dep}$ and the other instance-level depth loss $L^{I}_{dep}$:
\begin{align}
L_{dep} &= L^{P}_{dep}+\lambda^{I}_{dep} L^{I}_{dep},
\label{eq:total_depth_loss}
\end{align}
where $\lambda^{I}_{dep}$ controls the relative weighting and is set to 1 by default.
The pixel-level depth loss $L^{P}_{dep}$ calculates the depth error between depth predictions and ground truth inside each instance:
\begin{align}
L^{P}_{dep} &= \widetilde{L}_{dep}(D_{all}, \hat{D}_{all}).
\label{eq:pixel_depth_loss}
\end{align}
The instance-level depth loss $L^{I}_{dep}$ computes the depth error between instance depth shift $d^{s}$ and corresponding ground truth:
\begin{align}
L^{I}_{dep} &= \widetilde{L}_{dep}(d^{s},\hat{d}^{s}).
\label{eq:instance_depth_loss}
\end{align}
The minimum and average depth values within each instance mask are employed as ground truths of depth shift $d^{s}$ in Eq.~\ref{eq:t1} and Eq.~\ref{eq:t2}, respectively.

%% file: Sec4_experiments.tex
\section{Experiments}
\label{sec:exp}
\subsection{Datasets}
We evaluate our method on Cityscapes~\cite{cordts2016the}, Cityscapes-DPS~\cite{qiao2021vip}, and SemKITTI-DPS~\cite{qiao2021vip} benchmarks.
Cityscapes is a challenging dataset for image segmentation. In this dataset, 5,000 images of a high resolution 1,024$\times$2,048 are annotated at the high-quality pixel level, and are divided into 2,975, 500, and 1,425 images for training, validation, and testing, respectively. Cityscapes panoptic segmentation benchmark evaluates 8 thing and 11 stuff classes.
Recently, Qiao \etal~\cite{qiao2021vip} proposed a depth-aware panoptic segmentation dataset Cityscapes-DPS by supplementing Cityscapes with depth annotations, which are computed from the disparity maps via stereo images. 2,400 and 300 images are annotated for training and validation, respectively.
SemKITTI-DPS~\cite{qiao2021vip} is converted from cloud points with semantic annotations in SemKITTI~\cite{behley2019semantickitti}, consisting of
19,130 and 4,071 images for training and validataion, respectively.
  
\subsection{Metrics}
\paragraph{Metric of PS.}
The results for panoptic segmentation are evaluated with the standard Panoptic Quality (PQ) metric, introduced by Kirillov et al.~\cite{panoptic}. The formulation of PQ is:
\begin{equation}
\mathrm{PQ}=\frac{\sum_{p,g\in TP}{\mathrm{IoU}(p,g)}}{|TP|+\frac{1}{2}|FP|+\frac{1}{2}|FN|},
\end{equation}
where $p$ and $g$ denote the predicted and ground truth segment, respectively. $TP$, $FN$, and $FP$ represent matched pairs of segments ($\mathrm{IoU}(p,g)>0.5$), unmatched ground truth segments, and unmatched predicted segments, respectively. PQ of both thing and stuff classes are reported.

\vspace{-4mm}
\paragraph{Metric of DPS.}
The evaluation metric for depth-aware panoptic segmentation is DPQ~\cite{qiao2021vip}, which quantifies the performance for segmentation and depth estimation simultaneously.
Specifically, given prediction $P$, ground-truth $\hat{P}$, and depth threshold $\lambda$ , $\mathrm{DPQ}^\lambda$ is computed as:
\begin{equation}
\mathrm{DPQ}^\lambda(P, \hat{P})=\mathrm{PQ}(P^{\lambda},\hat{P}).
\end{equation}
$P^{\lambda}=P$ for pixels that have absolute relative depth errors under $\lambda$ to filter out pixels that have large absolute relative depth errors.
DPQ is calculated by averaging $\mathrm{DPQ}^\lambda(P, \hat{P})$ over $\lambda = \{0.1, 0.25, 0.5\}$.

\begin{table*}
  \center
  \setlength{\tabcolsep}{5pt}
  \begin{tabular}{llc|ccc|ccc}
    \toprule
    Method&Backbone&Extra Data&PQ&PQ$^{\text{Th}}$&PQ$^{\text{St}}$&PQ [test]&PQ$^{\text{Th}}$ [test]&PQ$^{\text{St}}$ [test]\\
    \midrule
    UPSNet~\cite{xiong19upsnet}        &R-50&-&59.3&54.6&62.7&-&-&-\\
    Seamless~\cite{porzi2019seamless}      &R-50&-&59.8&54.6&63.6&-&-&-\\
    UPSNet~\cite{xiong19upsnet}        &R-50&COCO&60.5&57.0&63.0&-&-&-\\
    SSAP$\dagger$~\cite{SSAP_Gao_ICCV}             &R-101&-&61.1&55.0&-   &58.9&48.4&66.5\\
    Unifying~\cite{Li_2020_CVPR}          &R-50&-&61.4&54.7&66.3&61.0&52.7&67.1\\
    Panoptic-DeepLab~\cite{cheng2020panoptic}&Xcp-71~\cite{chollet2017xception:}&-&63.0&-&- &- &-&-\\
    Panoptic-DeepLab$\dagger$&Xcp-71&-&-&-&- &\textbf{62.3}&52.1&\textbf{69.7}\\
    \midrule
    Baseline, original~\cite{li2021fully}       &R-50&-&61.4&54.8&66.6&-&-&-\\
    Baseline, our impl.  &R-50&-&62.4&56.0&67.1&-&-&-\\
     + AKF    &R-50&-&63.6&57.2&68.1&-&-&-\\
     + FSF    &R-50&-&\textbf{64.1}&\textbf{58.8}&\textbf{68.1}&62.0&\textbf{55.0}&67.1\\
    \bottomrule
  \end{tabular}
  \caption{Panoptic segmentation results on Cityscapes validation and test sets. `AKF': adaptive kernel fusion. `FSF': full-scale fine-tuning. $\dagger$: test-time augmentation. Results are reported as percentages.}
  \label{tab:ps}
  \setlength{\tabcolsep}{3.5pt}
  \center
  \begin{tabular}{llc|c|c|c|c}
    \toprule
    Method&Backbone&Extra Data&{$\lambda$=0.5}&{$\lambda$=0.25}&{$\lambda$=0.1}&DPQ\\
    \midrule
    ViP-DeepLab$\dagger$&WR-41~\cite{chen2020scaling}&MV, CSV&68.7 / 61.4 / 74.0&66.5 / 60.4 / 71.0&50.5 / 45.8 / 53.9&\textbf{61.9} / 55.9 / \textbf{66.3}  \\
    Ours&R-50&-&65.6 / 59.2 / 70.2&62.3 / 57.0 / 66.1&43.2 / 40.7 / 45.1&57.0 / 52.3 / 60.5 \\
    Ours&Swin-T~\cite{liu2021swin}&-&66.5 / 61.0 / 70.5&64.1 / 59.9 / 67.2&48.6 / 44.8 / 51.3&59.7 / 55.2 / 63.0\\
    Ours&Swin-S~\cite{liu2021swin}&-&67.4 / 62.5 / 71.0&65.0 / 61.4 / 67.7&48.8 / 44.2 / 52.3&60.4 / \textbf{56.0} / 63.6\\
    \bottomrule
  \end{tabular}
  \caption{Depth-aware panoptic segmentation results on Cityscapes-DPS. `MV': Mapillary Vistas~\cite{neuhold2017mapillary}. `CSV': Cityscapes videos with pseudo labels~\cite{chen2020naive}. $\dagger$: test-time augmentation. Each cell contains DPQ / DPQ$^\text{Th}$ / DPQ$^\text{St}$ scores. Results are reported as percentages.}
  \label{tab:dps}
\end{table*}

\subsection{Implementation}
\label{sec:exp:imple}
Our models are implemented with PyTorch\cite{steiner2019pytorch:} and the Detectron2\cite{wu2019detectron2} toolbox. Unless specified, ResNet-50~\cite{he2016deep} with FPN~\cite{lin2017feature} is employed as backbone. 
The dimension of mask embeddings $E^m$ is set to 256 following PanopticFCN~\cite{li2021fully}. The dimension of depth embeddings $E^d$ is set to 16.

The training details of PS and DPS model are introduced below, where the trained PS model is used to initialize the DPS model.

\vspace{-4mm}
\paragraph{Training details of PS model.}
The training of panoptic segmentation model consists of two steps, where the first step takes a large mini-batch of small cropped images, while the second one has a small mini-batch of large full-scale samples.
Specifically, we first train PanopticFCN model with Adam~\cite{kingma2014adam} for 130k iterations using synchronized batch normalization~\cite{ioffe2015batch}.
The learning rate is initialized as $10^{-4}$ and the \emph{poly} schedule with power 0.9 is adopted. 
The filters in the first two stages of the backbone are fixed in Detectron2 by default, but we find updating these parameters can boost around 0.5\% PQ.
The images are resized with random factors in [0.5, 2.0], and then cropped into 512$\times$1024. Each mini-batch contains 32 samples.
Color augmentation~\cite{liu2016ssd:} and horizontal flipping are employed during training.
At the second step, we fine-tune the PS model with images scaled by [1.0, 1.5] and then cropped into 1024$\times$2048 for 10k iterations. The batch size is 8. BN layers and the first two stage layers of the backbone are fixed in this and following training steps.

\vspace{-4mm}
\paragraph{Training details of DPS model.}
We train the entire model on Cityscapes-DPS for 10k iterations.
Images are resized by [0.8, 1.2] and depth annotations are scaled accordingly.
We put 8 samples in one mini-batch, which are center-cropped into 1024$\times$2048.
Color augmentation and horizontal flipping are also employed.
On SemKITTI-DPS, we use the pre-trained model from Cityscapes, but the training samples are cropped into 384$\times$1280 at both pre-training and fine-tuning steps. Other settings remain the same. 

The total training loss is:
\begin{equation}
L = \lambda_{pos} L_{pos}+ \lambda_{seg} L_{seg} + \lambda_{dep} L_{dep}.
\end{equation}
$L_{dep}$ is defined in Eq.~\ref{eq:total_depth_loss}. $L_{pos}$ and $L_{seg}$ are losses for classification and segmentation separately as described in PanopticFCN~\cite{li2021fully}.  $\lambda_{pos}$, $\lambda_{seg}$, and $\lambda_{dep}$ are set to 1, 4, and 5, respectively. 
Training steps for PS and DPS model tasks 46 and 5 hours, respectively, on 8 Nvidia 2080Ti GPUs.

The predictions of our method are obtained by a single model and a single inference. No test-time augmentations such as horizontal flipping or multi-scale testing are adopted.

\begin{table*}[t]
  \setlength{\tabcolsep}{6.5pt}
  \center
  \begin{tabular}{c|ccc|c|c|c|c}
    \toprule
    Variants&IDE&IDN&$L^{I}_{dep}$&$\lambda$=0.5&$\lambda$=0.25&$\lambda$=0.1&DPQ\\
    \midrule
    A&&&                            &64.0 / 56.8 / 69.0&60.0 / 52.8 / 64.8&40.8 / 35.9 / 44.9&54.9 / 48.5 / 59.6\\
    B&\checkmark& &   &63.8 / 56.4 / 69.2&60.0 / 52.6 / 65.4&41.0 / 34.1 / 46.0&54.9 / 47.7 / 60.2\\
    C&\checkmark&$\mathcal{T}_1$&   &64.9 / 57.1 / 70.2&61.0 / 53.2 / 66.2&42.4 / 36.5 / 45.5&56.1 / 48.9 / \textbf{60.6}\\
    D&\checkmark&$\mathcal{T}_2$&   &65.0 / 57.3 / 70.1&60.9 / 53.2 / 66.2&42.5 / 36.7 / 45.3&56.1 / 49.1 / 60.5\\
    E&\checkmark&$\mathcal{T}_1$&\checkmark&60.0 / 47.2 / 69.2&49.5 / 34.8 / 60.1&29.1 / 21.3 / 34.8&46.2 / 34.4 / 54.7\\
    F&\checkmark&$\mathcal{T}_2$&\checkmark&65.6 / 59.2 / 70.2&62.3 / 57.0 / 66.1&43.2 / 40.7 / 45.1&\textbf{57.0} / \textbf{52.3} / 60.5\\
    \bottomrule
  \end{tabular}
  \vspace{-.2em}
  \caption{Ablation studies on Cityscapes-DPS. `IDE':  instance-wise depth estimation. `IDN': instance depth normalization. Each cell contains DPQ / DPQ$^\text{Th}$ / DPQ$^\text{St}$ scores. Results are reported as percentages.}
  \label{tab:dps_ablation}
  \vspace{-.3em}
\end{table*}
\begin{table}
  \center
  \begin{tabular}{c|c|c}
    \toprule
    Methods&RMSE [val]$\downarrow$ &RMSE [test]$\downarrow$\\
    \midrule
	MGNet$^{\ddagger}$~\cite{schon2021mgnet}&8.30&-\\
	Laina \etal~\cite{laina2016deeper} &-&7.27\\
	Pad-Net~\cite{xu2018pad} &-&7.12 \\
	Zhang \etal~\cite{zhang2018joint}&-&7.10\\
	SDC-Depth$^{\ddagger}$~\cite{wang2020sdc}&-&6.92\\
	Ours$^{\ddagger}$&\textbf{6.91}&\textbf{6.69}\\
    \bottomrule
  \end{tabular}
   \vspace{-.2em}
  \caption{Monocular depth estimation results on Cityscapes. $\ddagger$: methods utilize panoptic segmentation annotations.}
  \label{tab:depth}
\end{table}
\begin{table}
  \center
  \setlength{\tabcolsep}{4.5pt}
  \begin{tabular}{c|ccc|ccc}
    \toprule
    Variants& IDE&IDN&$L^{I}_{dep}$&DPQ&DPQ$^\text{Th}$&DPQ$^\text{St}$\\
    \midrule
    A&&&                                   &45.3&41.4&47.5\\
    B&\checkmark& &          &45.8&42.9&47.5\\
    C&\checkmark&$\mathcal{T}_1$&          &46.7&45.6&{\bf 47.6}\\
    D&\checkmark&$\mathcal{T}_2$&          &46.7&45.5&{\bf 47.6}\\
    E&\checkmark&$\mathcal{T}_1$&\checkmark&36.1&35.1&40.6\\
    F&\checkmark&$\mathcal{T}_2$&\checkmark&{\bf 46.9}&{\bf 46.0}&{\bf 47.6}\\
    \bottomrule
  \end{tabular}
   \vspace{-.2em}
  \caption{Ablation studies on SemKITTI-DPS. `IDE': instance-wise depth estimation. `IDN': instance depth normalization. Results are reported as percentages.}
   \vspace{-.2em}
  \label{tab:kitti}
\end{table}

\subsection{Main Results}
\paragraph{Panoptic segmentation.}
The original PanopticFCN~\cite{li2021fully} achieves 61.4\% PQ on the Cityscapes validation set, whereas our implementation achieves 62.4\% PQ without introducing extra parameters. 
With the adaptive kernel fusion mechanism and the full-scale fine-tuning step, our full model achieves 64.1\% PQ. 
Because the test set annotations are withheld, we submit the test set results to the online evaluation server. Results are shown in Table~\ref{tab:ps}. Our method achieves the highest 55.0\%  PQ$^{\text{Th}}$.

\vspace{-4mm}
\paragraph{Depth-aware panoptic segmentation.}
In Table~\ref{tab:dps}, we compare our method with ViP-DeepLab, which is the only published work for depth-aware panoptic segmentation. It achieves a higher performance, however, please note that extra engineering tricks including pre-training with larger datasets~\cite{neuhold2017mapillary}, semi-supervised learning~\cite{chen2020naive}, AutoAug~\cite{cubuk2019autoaugment}, and test-time augmentation are employed in ViP-DeepLab, which can also be utilized in our method to further improve the performance.

\vspace{-4mm}
\paragraph{Monocular depth estimation.}
To further demonstrate the superiority of our method over SOTA methods, we report the results of the monocular depth estimation on Cityscapes. As shown in Table~\ref{tab:depth}, the proposed approach shows a clear advantage even compared to methods that boost depth estimation with panoptic segmenation annotations.

\begin{figure*}
  \center
  \includegraphics[width=1.\linewidth]{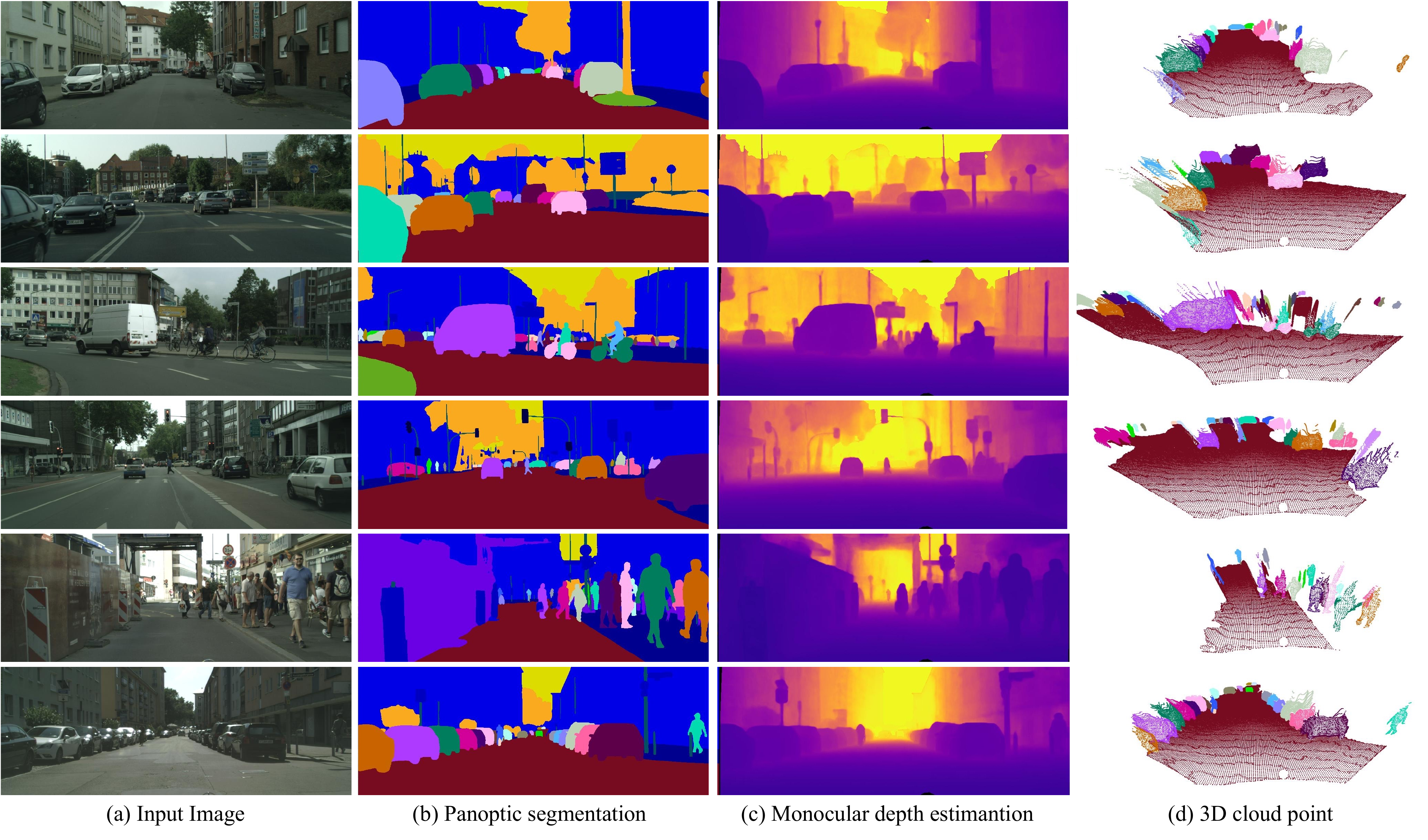}
  \vspace{-1.5em}
  \caption{Example predictions of our model.
    The panoptic segmentation and monocular depth estimation results are predicted from the input image with a unified model. 
    Best viewed in color and zoom.
    }
  \label{fig:vis_more}
\end{figure*}

\vspace{-4mm}
\paragraph{Visualization results.}
The panoptic segmentation, monocular depth estimation, and corresponding 3D point cloud results can be visualized in Figure~\ref{fig:vis_more}. 

\begin{figure}
  \center
  \includegraphics[width=1.\linewidth]{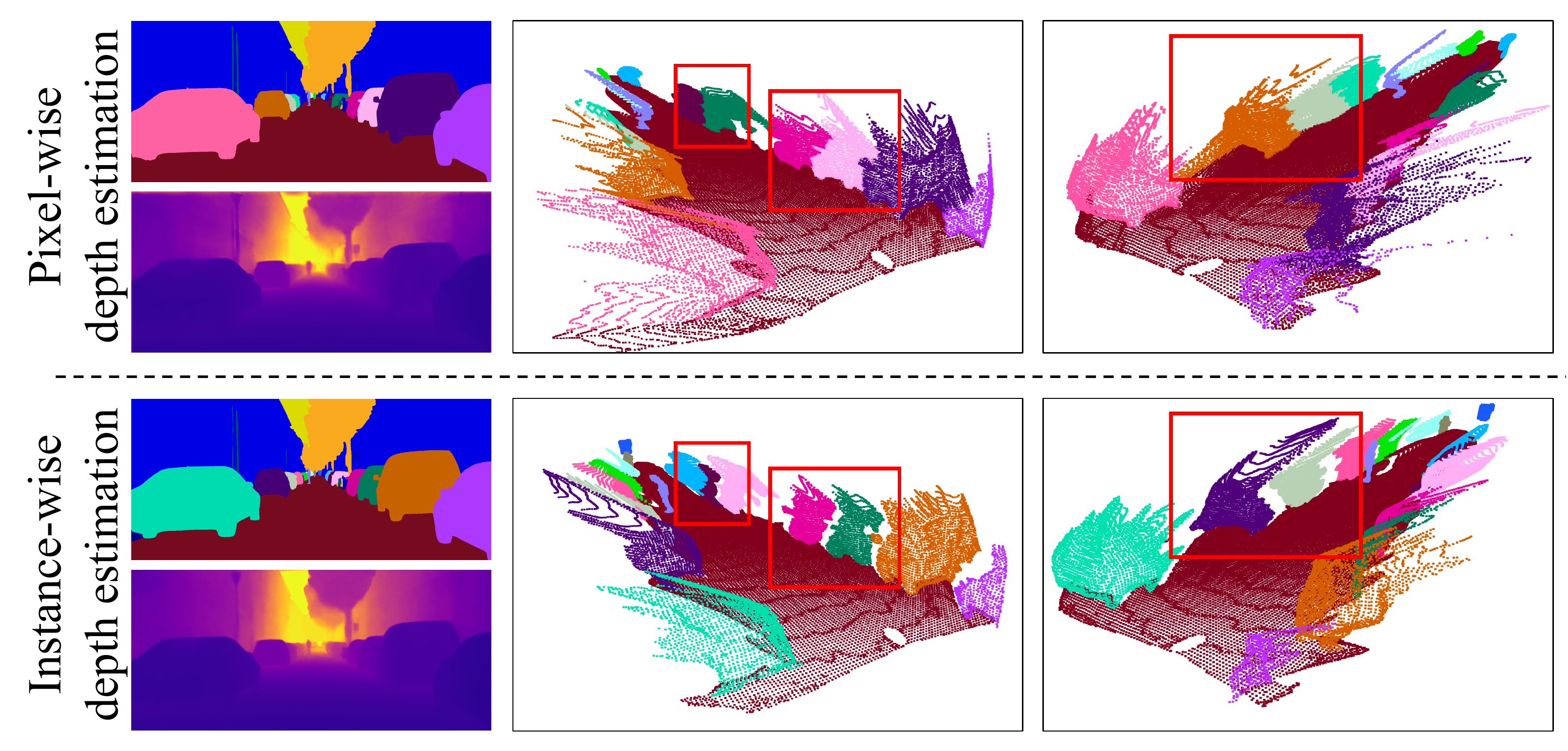}
  \caption{Pixel-wise depth estimation outputs smooth values at the boundary of two instances, whereas instance-wise depth estimation can generate more reasonable discontinuous depth values.
  }
  \label{fig:vis_boundary}
\end{figure}
\subsection{Ablation Experiments}
\paragraph{Influence of adaptive kernel fusion.}
Unlike PanopticFCN that generates stuff kernels by global average pooling the masked kernel map, our approach adaptively fuses stuff kernels (Eq.~\ref{eq:akf}), avoiding the interference of varying stuff region sizes.
As shown in Table~\ref{tab:ps}, employing adaptive kernel fusion (AKF) mechanism improves PQ$^{\text{Th}}$ and PQ$^{\text{St}}$ by 1.2\% and 1.0\%, respectively. 

\vspace{-4mm}
\paragraph{Full-scale fine-tuning.}
Fine-tuning the learned PS model with full-scale images (FSF) promotes the ability to distinguish distant instances.
As shown in Table~\ref{tab:ps}, FSF further boosts PQ$^{\text{Th}}$ by 1.6\% to 64.1\% PQ. 

\vspace{-4mm}
\paragraph{Instance-wise depth estimation.}
The comparison of pixel- and instance-wise depth estimation is quantified with variant-A and -B in Table~\ref{tab:dps_ablation}.
Variant-A performs pixel-wise depth regression by setting the dimension $e^d_2$ of the depth embedding map $E^d$ to 1, achieving 54.9\% DPQ.
Variant-B conducts instance-wise depth estimation with dynamic kernels, which achieves similar performance to variant-A.
However, as shown in Figure~\ref{fig:vis_boundary}, pixel-wise depth estimation tends to output smooth values at object boundaries. By contrast, our instance-wise depth prediction method avoids this and can generate more reasonable discontinuous depth values at the boundary of two instances.

\vspace{-4mm}
\paragraph{Instance depth normalization.}
In Table~\ref{tab:dps_ablation}, variant-C and -D further normalizes the instance depth maps with depth shift and depth range respectively, as described in Eq.~\ref{eq:t1} and Eq.~\ref{eq:t2}. 
These two variants achieve similar performances but are about 1.2\% DPQ higher than variant-A and -B.
This result demonstrates the effectiveness of leveraging instance depth statistics for depth estimation.

\vspace{-4mm}
\paragraph{Instance-level depth loss.}
As shown in Table~\ref{tab:dps_ablation}, variant-E and -F employs the instance-level depth loss upon variant-C and -D, in which the depth shift $d^{s}$ is supervised with the minimum or mean depth value within each instance mask.
Although variant-E drops the performance, variant-F significantly boosts DPQ$^{\text{Th}}$ by 3.2\%.
This is because the ground truth depth is noisy, and minimum depth values are sensitive to noise while mean depth values not.

\vspace{-4mm}
\paragraph{Results on SemKITTI-DPS.}
Table~\ref{tab:kitti} shows our proposed instance-wise depth estimation (variant-F) achieves better performance again on the dataset of SemKITTI-DPS, especially for thing instances (46.0\% v.s. 41.4\%  DPQ$^{\text{Th}}$), which is consistent with the results on Cityscapes-DPS (Table~\ref{tab:dps_ablation}).